\def\year{2022}\relax
\documentclass[letterpaper]{article} 
\usepackage{aaai22}  
\usepackage{times}  
\usepackage{helvet}  
\usepackage{courier}  
\usepackage[hyphens]{url}  
\usepackage{graphicx} 
\usepackage{natbib}  
\usepackage{caption} 
\DeclareCaptionStyle{ruled}{labelfont=normalfont,labelsep=colon,strut=off} 
\usepackage[switch]{lineno}

\usepackage{algorithm}
\usepackage{algorithmic}
\usepackage{booktabs}
\usepackage{multirow}
\usepackage{amssymb}
\usepackage{subcaption}

\usepackage{newfloat}
\usepackage{listings}
\floatstyle{ruled}
\newfloat{listing}{tb}{lst}{}
\floatname{listing}{Listing}
\title{Deepfake Network Architecture Attribution}
\author{
	Tianyun Yang$^{1,2}$\equalcontrib, Ziyao Huang$^{1,2}$\equalcontrib, Juan Cao$^{1,2}$\thanks{Corresponding author}, Lei Li$^{1,2}$, Xirong Li$^{3}$
}

\affiliations{
	$^1$ Key Lab of Intelligent Information Processing, Institute of Computing Technology, CAS, Beijing, China\\
    $^2$ University of Chinese Academy of Sciences, Beijing, China\\
    $^3$ Key Lab of Data Engineering and Knowledge Engineering, Renmin University of China\\
    \{yangtianyun19z,huangziyao19f,caojuan,lilei17b\}@ict.ac.cn, xirong@ruc.edu.cn
}

\usepackage{bibentry}
\usepackage{hyperref}

\usepackage{xcolor}

\begin{document}

\maketitle

\begin{figure*}
\begin{center}
\includegraphics[width=\linewidth]{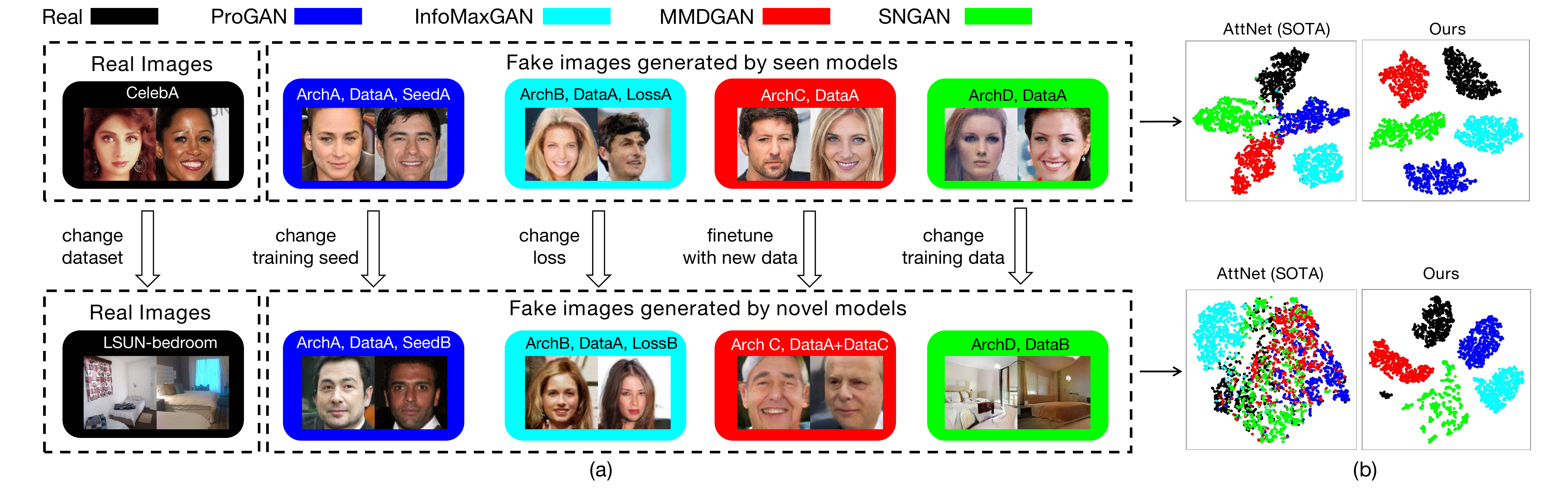} 
\end{center}
  \caption{(a) The scenario for deepfake network architecture attribution. (b) The t-SNE visual comparison between our learned features and AttNet~\cite{yu2019attributing}. When testing on images from the same set of GAN models and real images used in training (above), AttNet and our method both extract distinct features. However, when testing on novel images from finetuned models or models with changed seed, loss or dataset (below), features extracted by AttNet are highly entangled, but our method can still extract well-separated feature.}   
\label{fig:intro}
\end{figure*}

\begin{abstract}


With the rapid progress of generation technology, it has become necessary to attribute the origin of fake images. 
Existing works on fake image attribution perform multi-class classification on several Generative Adversarial Network (GAN) models and obtain high accuracies. While encouraging, these works are restricted to model-level attribution, only capable of handling images generated by seen models with a specific seed, loss and dataset, which is limited in real-world scenarios when fake images may be generated by
privately trained models. This motivates us to ask whether it is possible to attribute fake images to the source models' architectures even if they are finetuned or retrained under different configurations. In this work, we present the first study on \textit{Deepfake Network Architecture Attribution} to attribute fake images on architecture-level. 
Based on an observation that GAN architecture is likely to leave globally consistent fingerprints while traces left by model weights vary in different regions, we provide a simple yet effective solution named DNA-Det for this problem. Extensive experiments on multiple cross-test setups and a large-scale dataset demonstrate the effectiveness of DNA-Det. Our source code and dataset can be found here:
\url{https://github.com/ICTMCG/DNA-Det}




\end{abstract}
 

\section{Introduction}

The deepfake technology has raised big challenges to visual forensics. Dedicated research efforts are paid~\cite{durall2020watch,wang2020cnn,liu2020global,zhang2019detecting,
	jeon2020t,nataraj2019detecting,chai2020makes,frank2020leveraging,zhao2021multi,haliassos2021lips,liu2021spatial,chandrasegaran2021closer,li2020identification} to detect generated images in recent years. However, only real/fake classification is not the end: On the one hand, for malicious and illegal content, law enforcers need to identify its owner. On the other hand, GAN models need experienced designers with laborious trial-and-error testings, some of which have high commercial value and should be protected. These motivate works on fake image attribution, i.e., attributing the origin of fake images. For fake image attribution, existing works~\cite{marra2019gans,yu2019attributing,frank2020leveraging,joslin2020attributing} perform attribution for multiple GAN models and obtain high classification accuracies. While encouraging, the problem of GAN attribution is far from studied and solved sufficiently. 

From the perspective of understanding GAN fingerprints, previous works~\cite{marra2019gans,yu2019attributing,frank2020leveraging,joslin2020attributing} suggest that: 1) Models with different architectures have distinct fingerprints. 2) With architecture fixed, changing only the model's random initialization seed or training data also results in a distinct fingerprint. From 2), it can be deduced that model weights may influence GAN fingerprints. While from 1), it cannot be verified whether the GAN fingerprint is related to the architecture since weights also change as the architecture changes. This motivates us to investigate whether GAN architectures leave fingerprints. In other words, do different models with the same architecture share the same fingerprint? Answering this question may help us understand deeper into the generation of GAN fingerprints.

From the perspective of application, previous works on GAN attribution only perform model-level attribution, i.e., training and testing images come from the same model, which means for fake images, we can only handle those generated by seen models. However, this approach is limited in real-world scenarios. For malicious content supervision, the malicious producers would probably download a certain deepfake project to their own computers from code hosting platforms at first, and then use their personal collected training data to finetune or train from scratch instead of directly using the public available models. In such a situation, model-level attribution is no longer applicable since it is unfeasible to get the privately trained model. For intellectual property protection, if an attacker steals a copyrighted GAN, and modifies weights by finetuning, model-level attribution will fail too. These motivate us to solve fake image attribution under a more generic setting, i.e., attribute fake images to the source architecture instead of the specific model.

\begin{figure}[t]
\begin{center}
\includegraphics[width=\linewidth]{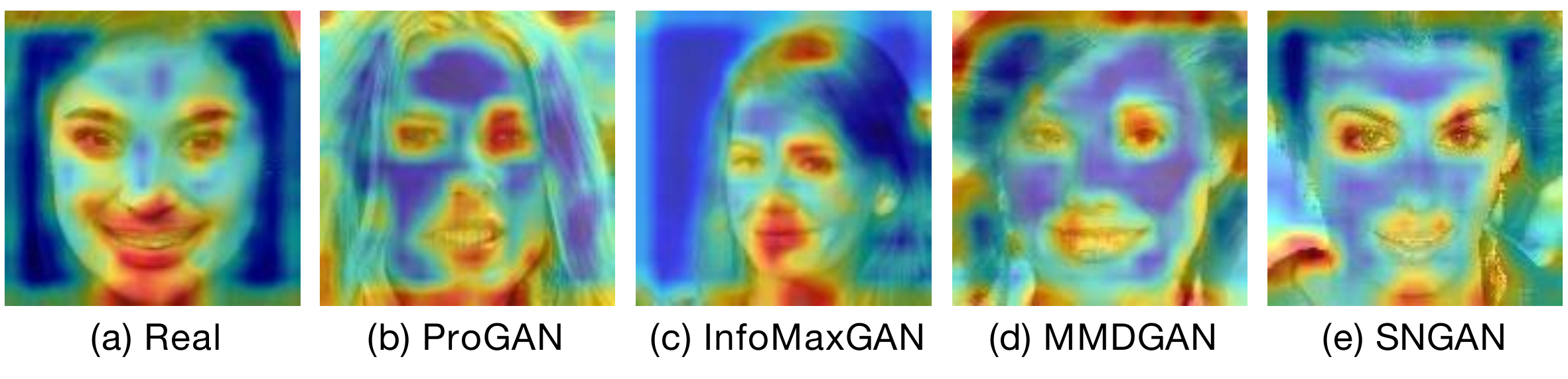} 
\end{center}
  \caption{Class activation maps from trained AttNet classifying \{real, ProGAN, InfoMaxGAN, MMDGAN, SNGAN\}.} 
\label{fig:gradcam}
\end{figure}


In this paper, we propose a novel task of \textbf{Deepfake Network Architecture
Attribution}. Compared with model-level attribution, architecture-level attribution 
requires attributing fake images to their generators' architectures even if the models are fine-tuned or retrained with a different seed, loss or dataset. Although architecture-level attribution is more coarse-grained than model-level attribution, it is still challenging. 
As Figure~\ref{fig:intro} shows, for seen GAN models with certain architectures (above), there may exist other versions of novel models different in training seed, loss or training data (below). If we train on real images and seen models, as the t-SNE plots show, although AttNet~\cite{yu2019attributing} extracts distinct features when testing on images generated by seen models (above), features are highly entangled on images from novel models (below). 
To explain the drop, we visualize what regions the network focuses on for attribution in Figure~\ref{fig:gradcam}. We notice that the network tends to focus on local regions closely related to image semantics such as eyes and mouth. However, for architecture attribution, it is problematic to concentrate on semantic-related local regions. 


In this work, we observe that: GAN architecture is likely to leave fingerprints, which are globally consistent among the full image instead of gathered in local regions. Besides, traces left by weights varies in different regions. This observation is based on an empirical study in Section~\ref{sec:study}. Specifically, we divide GAN images into patches of equal size and conduct model weight classification and architecture classification on patches. We train on patches from a single position and then test on patches from every position respectively.
We can observe that:
1) In weight classification, the testing accuracy is high on patches with the same position as patches used in training, but drops largely on patches from other positions. This result indicates that traces left by model weights are likely associated with the position. 2) In architecture classification, testing accuracies on patches from all positions are higher than 90$\%$, even though we trained solely on patches from a single position. This suggests that there exist globally consistent traces on GAN images, which are distinct for models of different architectures. This globally consistent distinction is probably caused by the architecture under the prior observation from 1) that weight traces vary in different regions. 


Motivated by the observation above, it is foreseeable that if we concentrate on globally consistent traces, architecture traces would play a primary role in decision, which generalize better when testing on unseen models. Thus we design a method for GAN architecture attribution, which we call DNA-Det: \underline{D}eepfake \underline{N}etwork \underline{A}rchitecture \underline{Det}ector. DNA-Det explores globally consistent features that are invariant to semantics to represent GAN architectures by two techniques, i.e., pre-training on image transformation classification and patchwise contrastive learning. The former helps the network to focus on architecture-related traces, and the latter strengthen the global consistency of extracted features.


To summarize, the contributions of this work include:
\begin{itemize}
\setlength{\itemsep}{0pt}
\setlength{\parsep}{0pt}
\setlength{\parskip}{0pt}
\item 
We propose a novel task of \textit{Deepfake Network Architecture Attribution} to attribute fake images to the source architectures even if models generating them are finetuned or retrained with a different seed, loss function or dataset. 
\item 
We develop a simple yet effective approach named DNA-Det to extract architecture traces, which adopts pre-training on image transformation classification and patchwise contrastive learning to capture globally consistent features that are invariant to semantics.
\item 
The evaluations on multiple cross-test setups and a large-scale dataset verify the effectiveness of DNA-Det. DNA-Det maintains a significantly higher accuracy than existing methods in cross-seed, cross-loss, cross-finetune and cross-dataset settings.

\end{itemize}

\begin{figure*}
\begin{center}
\includegraphics[width=\textwidth,height=3.8cm]{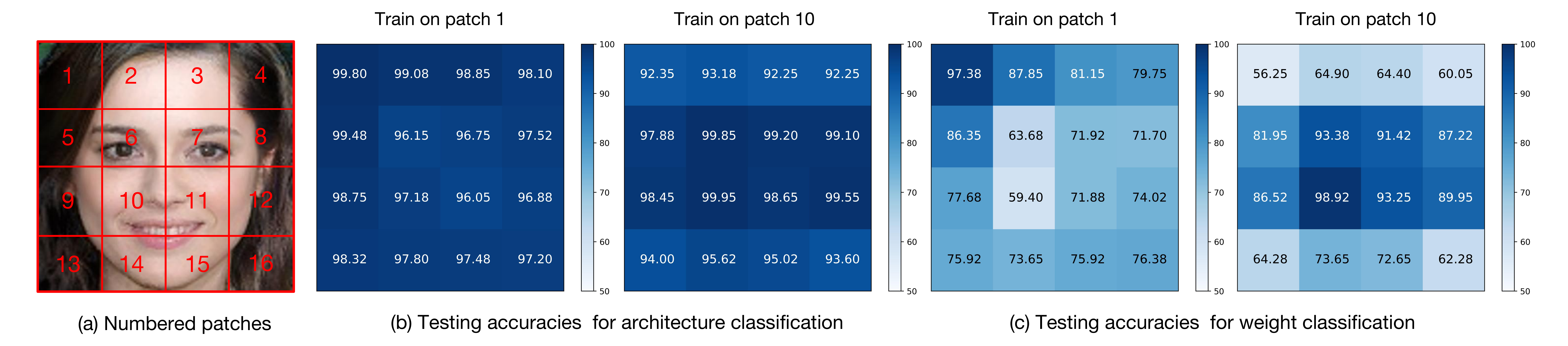} 
\end{center}
   \caption{
  Empirical study on GAN fingerprint in architecture and weight classification. We only train on patches from a fixed position, and test on patches from all positions respectively. We show the results training on patch 1 and 10.
}   
\label{fig:patch_acc}
\end{figure*}

\section{Related Work}


\textbf{Fake image attribution} can be classified into \textit{positive  attribution}~\cite{kim2020decentralized,yu2020artificial,yu2020responsible} and \textit{passive attribution}~\cite{marra2019gans,yu2019attributing,frank2020leveraging,joslin2020attributing,xuan2019scalable}. This paper focuses on passive attribution. Works on positive attribution insert artificial fingerprint~\cite{yu2020artificial,yu2020responsible} or key ~\cite{kim2020decentralized} directly into the generative model. Then when tracing the source model, the fingerprint or key can be decoupled from generated images. Positive attribution requires ``white-box" model training and thus is limited in ``black-box" scenario when only generated images are available. Passive attribution aims at finding the intrinsic differences between different types of generated images without getting access to the generative model, which is more efficient and challenging compared with positive attribution. The work in~\cite{marra2019gans} finds averaged noise residual can represent GAN fingerprint. The work in~\cite{yu2019attributing} decouples GAN fingerprint into model fingerprint and image fingerprint. Specifically, this work takes the final classifier features and reconstruction residual as the image fingerprint and the corresponding classifier parameters in the last layer as the model fingerprint. The work in~\cite{frank2020leveraging} observes the discrepant DCT frequency spectrums exhibited by images generated from different GAN architectures, and then sends the DCT frequency spectrum into classifiers for source identification. The work in~\cite{joslin2020attributing} derive a similarity metric on the frequency domain for GAN attribution. Above works on passive fake image attribution all conduct experiments on multiple GAN models and achieve high accuracy. While encouraging, these works are restricted to model-level attribution (i.e., training and testing images come from the same set of models), which is limited in the real scenario. In this paper, we propose to solve fake image attribution on architecture-level, which can attribute generative models to their architectures even if they are modified by fine-tuning or retraining. 

\noindent{\textbf{Pre-taining on image transformations}} was previously used in image manipulation detection~\cite{wu2019mantra,huh2018fighting} based on the assumption that there may exist post-processing discontinuity between the tampered region and its surrounding. Our approach is inspired by these works but driven by a different motivation: Many traditional image transformation functions are similar to image generation operations, and pre-training on classifying different image transformations can help the network focus on architecture-related globally consistent traces.


\begin{figure*}
\begin{center}
\includegraphics[width=1\textwidth]{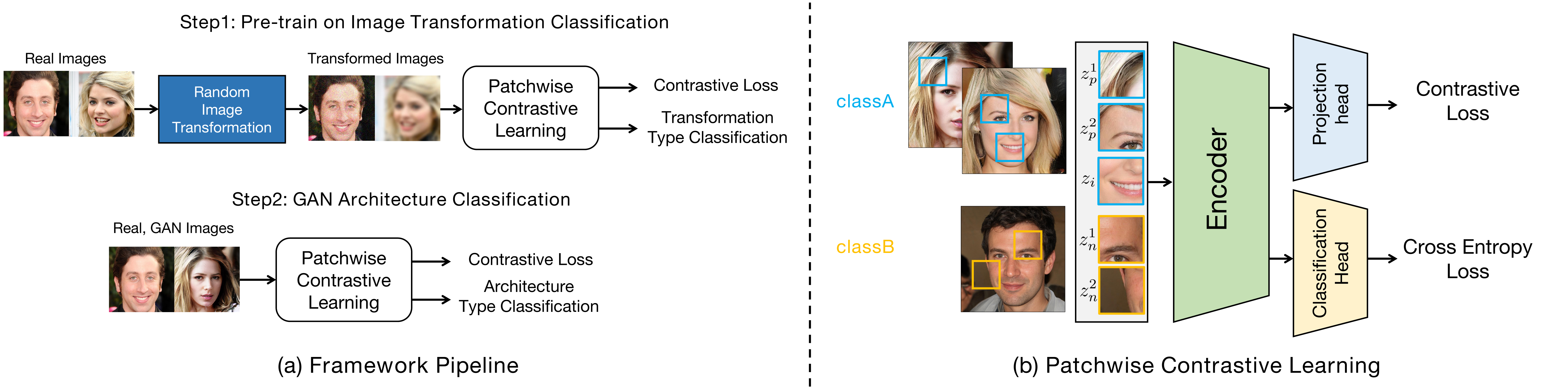} 
\end{center}
   \caption{{\bf Overview of DNA-Det's learning pipeline.} (a) Framework pipeline. In the first step, we use image-transformation classification as a self supervision task to make the network focus on architecture-related traces. In the second step, we use the weights learned in the former step as the initial weight and conduct GAN architecture classification. In the two steps, the network is trained by a patchwise contrastive learning mechanism. (b) Patchwise contrastive learning used in the two steps in (a), which force patches of the same class (image transformation type or GAN architecture type) close-by, and patches of different classes far apart.}   
\label{fig:method}
\end{figure*}

\section{Empirical Study on GAN Fingerprint}
\label{sec:study}
Reviewing fake images' generation process, the network components (e.g. convolution, upsampling, activation, normalization and so on) all operate on feature maps spatially equal, thus traces left by these components is likely to be identical on every patch. Intuitively, we hypothesize that if GAN architecture leaves traces, they would be globally consistent among patches. 
To verify this hypothesis, we design an empirical study as follows: We conduct two attribution experiments: 1) Architecture classification. We do four-class classification classifying images from four GAN models with different architectures, including ProGAN~\cite{karras2017progressive}, MMDGAN~\cite{binkowski2018MMD}, SNGAN~\cite{miyato2018spectral} and InfoMaxGAN~\cite{lee2021infomax}, all of which are trained on celebA dataset~\cite{liu2015celeba}. 2) Weight classification. Another four-class classification classifying images from four ProGAN models trained on celebA but with different training seed. It is foreseeable that in weight classification, the network will depend on weight traces for classification. In architecture classification, the network may depend on architecture or weight traces, or both. Implementation details are provided in the Appendix.

In detail, as Figure~\ref{fig:patch_acc}(a) shows, we divide each image into 4$\times$4 grid of patches and number them from 1 to 16 according to the position. We only train on patches from a fixed position and test on patches from all positions respectively, getting 16 testing accuracies for 16 positions.
Fig.~\ref{fig:patch_acc}(b)(c) shows the testing accuracies in architecture and weight classification when train solely on patches from position 1 and position 10. From the experiment results, we have two observations: 1) Weight traces is likely associated with the position, a semantic association perhaps. Since in weight classification, when the network is trained on patches from a single position, the testing accuracy is high on this position, but drops a lot on patches from other positions. 2) Architecture is likely to leave fingerprints, which are globally consistent among the full image. In architecture classification, testing accuracies on all positions are higher than 90$\%$ even though only patches from one single position are used for training. Given the prior observation from 1) that weight traces varies in different regions, this globally consistent distinction is probably caused by the architecture.


Our goal is to get a stable architecture representation regardless of weights. Given the empirical observations above, an intuitive approach is to restrict the global consistency of extracted features, such that architecture traces would play a decisive role in conducting architecture attribution.

\section{Proposed Approach}
\noindent{\textbf{Problem definition.}}  We set deepfake network architecture attribution as a multi-class classification problem. Given an image $x^{y}$ with source $y\in\mathbb{Y}=\{real,G_{1},G_{2},\dots,G_{N}\}$, where $G_{1},\dots,$ $G_{N}$ are different architectures. Our goal is to learn a mapping $D(x^{y}) \rightarrow y$. Note that the architecture in this paper refers to the architecture of the generator. Loss functions and discriminator are not considered as part of the architecture
, since they only influence the generator's weights indirectly by gradient back propagation, while our goal is to attribute fake images to the source architecture regardless of model weights.

\noindent{\textbf{Framework overview.}} Figure~\ref{fig:method} overviews the learning pipeline for DNA-Det. We train our network by two steps. In the first step, we use image transformation classification as a pre-train task to make the network focus on architecture-related traces. In the second step, we use the weights learned in former step as the initial weight and conduct GAN architecture classification. In the two steps, we use patchwise contrastive learning to force patches of the same class (image transformation type or GAN architecture type) close-by, and patches of different classes would be pushed far apart. 

\subsection{Pre-train on Image Transformations}
\label{sec:pretrain}
Given a certain number of GAN images with architecture labels, the obvious idea is to use these labels to train a classifier using a supervised objective such as cross-entropy loss. However, directly using features learned by supervised training is problematic. 
This would make the classifier harvest any useful features to help classification, which may include semantic-related information as shown in Figure~\ref{fig:gradcam}.
Inspired by works in~\cite{huh2018fighting, wu2019mantra}, we use image transformation classification as a pre-train task motivated two reasons: 1) We found that some traditional image transformation operations are similar to the generator's components. For example, blurring and noising with kernels resembles convolution computation, and the resampling operation is similar to the upsampling layer. Thus traces left by traditional image transformations share similar properties with architecture traces. 2) Traditional image transformations and the generator's components both conduct on the images spatially equal. Thus pre-training on image transformation classification could aid the network to focus on globally consistent traces.

In detail, four image transformation families are considered: compression, blurring, resampling and noising.
We randomly choose the parameters for each operation from a discrete set of numbers. Each operation with a specific parameter is taken as a unique type and we finally get 170 types of image transformations. 
In training, we apply these transformations on a natural image dataset containing LSUN~\cite{yu2015lsun} and CelebA. Then we conduct patchwise contrastive learning (described in Section~\ref{sec:pcl}) to force patches with the same image transformation close-by and different image transformations far apart. We use this pre-trained model to initialize model weights.


\subsection{Patchwise Contrastive Learning}
\label{sec:pcl}

We adopt a contrastive learning mechanism on patches to strengthen the global consistency of extracted features. Details are shown in Figure~\ref{fig:method}(b). Instead of training on whole images, randomly cropped patches are used as input samples.
These patches are fed into an encoder followed by a projection head and a classification head.
The projection head consists of a two-layer MLP network, which maps representations to the space where a supervised contrastive loss~\cite{khosla2020supervised} is calculated. For an anchor patch, patches with the same class are positives, and patches with different classes are negatives. The contrastive loss forces patches from the same class closer in the representation space, and pushes patches from different classes farther away. 
Specifically, the contrastive loss is calculated as follows: 
\begin{equation}
L_{con}=\sum_{i \in I} \frac{-1}{P(i)} \sum_{p \in P(i)} \log \frac{\exp \left(z_{i} \cdot z_{p} / \tau\right)}{\sum_{a \in A(i)} \exp \left(z_{i} \cdot z_{a} / \tau\right)}
\end{equation}
Here, $i \in I$ is the index of an arbitrary training patch. $P(i)$ is the set of all positive pairs for the patch $i$. $A(i) \equiv I \backslash\{i\}$, which includes all positive and negative pairs for patch $i$. $z_{i}$ is the feature vector for patch $i$. $z_{a}$ is the feature vector for patches in $A(i)$. $z_{p}$ and $z_{n}$ (shown in Figure~\ref{fig:method}(b)) refer to the feature vector for positive and negative pairs respectively. 


The classification head maps the representation from the encoder to the label space, in which we calculate a cross-entropy loss $L_{ce}$. Overall, the objective for patchwise contrastive learning is formulated as:
\begin{equation}
L=w_1 \cdot L_{con} + w_2 \cdot L_{ce}
\end{equation}
where $\omega_{1}$ and $\omega_{2}$ are non-negative weights. Automatic weighted learning mechanism~\cite{kendall2018multi} is used to adaptively optimize the objective.

\section{Experiments}

\subsection{Experimental Setup}

\noindent{\textbf{Compared Methods.}}
We compare our method with several representative methods for fake image attribution as follows: 1) PRNU~\cite{marra2019gans}: a method using photo-response non-uniformity (PRNU) patterns as the fingerprint for fake image attribution. 2) AttNet~\cite{yu2019attributing}: a PatchGAN-like classifier for fake image attribution. 3) LeveFreq~\cite{frank2020leveraging}: a frequency-based method that uses Discrete Cosine Transform (DCT) images for fake image attribution and detection.


\noindent{\textbf{Implementation Details.}}
 For the network architecture, we use a shallow 8-layer CNN network as the encoder. The output channel numbers for convolution layers are 64,64,128,128,256,256,512 and 512. A Global Average Pooling is added after the convolution layers. For patchwise contrastive learning, we firstly resize all images to 128px (the lowest resolution in the dataset), and then resize them to 512px to magnify GAN traces, on which we randomly crop 16 patches of 64px as inputs. For inference, we test on the full image instead of patches. For optimization, we choose Adam optimizer. For the celebA experiment in section~\ref{sec:q1}, the initial learning rate is set to $10^{-4}$ and is multiplied by 0.9 for every 500 iterations. For the LSUN-bedroom experiment in section~\ref{sec:q1} and the experiment in section~\ref{sec:q2}, the initial learning rate is set to $10^{-3}$ and is multiplied by 0.9 for every 2500 iterations. The batch size is 32 $\times$ \#classes in Section~\ref{sec:q1} and 16 $\times$ \#classes in Section~\ref{sec:q2} with a class balance strategy. For the GradCAM maps shown in this paper, we visualize on layer-4. More details of the experiments could be found in the appendix material. 

\begin{table}[t]
\begin{center}
\scalebox{0.6}{
\begin{tabular}{lccccc}
\toprule  
 Real, GAN & train-set  & cross-seed & cross-loss & cross-finetune & cross-dataset \\
\midrule  
\textbf{\textit{celebA}}\\
Real & celebA  & - & - & - & bedroom \\
ProGAN & celebA(seed0) & seed 1-9 & - & FFHQ-elders & bedroom\\
MMDGAN & celebA & -  & CramerGAN& FFHQ-elders& bedroom \\
SNGAN & celebA  & -  & - & FFHQ-elders&  bedroom \\
InfoMaxGAN & celebA & - & SSGAN & FFHQ-elders& bedroom\\
\midrule  
\textbf{\textit{LSUN-bedroom}}\\
Real & bedroom  & - & -& - & celebA \\
ProGAN & bedroom(seed0) & seed 1-9 & -& LSUN-sofa& celebA  \\
MMDGAN & bedroom & -  & CramerGAN& LSUN-sofa& celebA \\
SNGAN & bedroom  & -  & -& LSUN-sofa &  celebA \\
InfoMaxGAN & bedroom & - & SSGAN& LSUN-sofa & celebA \\
\bottomrule 
\end{tabular}}
\end{center}
\caption{
Dataset split for cross seed, loss, finetune and dataset evaluation. The evalution consists of two groups: celebaA and LSUN-bedroom.}
\label{tab:128att}
\end{table}

\begin{table*}
\begin{center}
\scalebox{0.73}{
\begin{tabular}{lcccccccccc}
\toprule
\multirow{2}{*}{Method} & \multicolumn{5}{c}{\textbf{\textit{celebA}}} & \multicolumn{5}{c}{\textbf{\textit{LSUN-bedroom}}}  \\
\cmidrule{2-6} \cmidrule{7-11} & closed-set  & cross-seed & cross-loss & cross-finetune & cross-dataset & closed-set & cross-seed & cross-loss & cross-finetune & cross-dataset\\
\midrule
PRNU(MIPR2019) & 90.64 &	20.69& 	29.33& 	48.88 	&21.27& 	66.31& 	30.50& 	35.23& 53.58& 	24.13 \\ 
LeveFreq(ICML20) & 99.50 &	86.02 	&92.50  &	52.19&	47.07 &	99.53& 	82.76 &76.30&74.59 &	53.42 \\ 
AttNet(ICCV19) & 98.88 &	83.50 &	87.10	& 35.21 &	38.54  &	99.25 &	88.73 &	89.28 &35.21 &	21.88\\ 
\midrule
AttNet+PT &99.50 &	93.46 &	82.65 &	36.89 &	44.71 &	98.77 &	98.19 &	97.60 &	73.29 &	49.82 \\ 
AttNet+PCL & \bf100.00 &	99.72 &	99.03 &	53.38 &	90.51 &\bf	100.00 &\bf	100.00 	&\bf100.00 &	95.69 &	81.06 \\ 
AttNet+PT+PCL &\bf 100.00 &	99.81 &	\bf99.80 &	57.48 &	93.76 &\bf	100.00 &	98.69 &	99.95 &	93.81 	&79.47\\ 
\midrule
Base & 99.88 &	94.17 &	62.93 &	45.83 &	33.02 & \bf	100.00 &72.73 &	74.83 & 	38.45 &	21.66  \\
Base+PT &\bf100.00 &	99.36 	&95.83&	54.80& 	53.00  & \bf 100.00 &	98.74 	&97.78 &61.74&	49.73  \\
Base+PCL &\bf100.00 &	99.96 &	98.30	& 77.89 &	89.93  &\bf 100.00 &	\bf 100.00 &	99.25 &	94.19 &	81.09 	 \\
\textbf{Base+PT+PCL(DNA-Det)} & \bf100.00 &	\bf99.99 &	99.53  &	\bf97.65&	\bf94.95 &	\bf100.00 &	99.99 &	99.90	 & \bf 97.50 &\bf 83.45\\  
\bottomrule
\end{tabular}}
\end{center}
\caption{
Evaluation on multiple cross-test setups for Section~\ref{sec:q1} and 
ablation study for Section~\ref{sec:q3} measured by accuracy. ``PT" means pre-train on image transformations. ``PCL" means patchwise contrastive learning.
} 
\label{tab:exp1}
\end{table*}

\subsection{Evaluation on Multiple Cross-Test Setups}
\label{sec:q1}
\noindent{\textbf{Datasets.}} This experiment is conducted on 5 classes: real, ProGAN, MMDGAN, SNGAN, InfoMaxGAN. Details of the dataset split are shown in Table~\ref{tab:128att}. As the table shows, the experiment is composed of two groups named by \textbf{\textit{celebA}} and \textbf{\textit{LSUN-bedroom}}, depending on the training dataset of the GAN models and real images in the train-set. For each group, we conduct cross-seed, cross-loss, cross-finetune, and cross-dataset testings to evaluate the generalization of architecture attribution on unseen models with different random seed, loss function and dataset from the models in the train-set. Specifically, for cross-seed testing, the ProGAN model in the train-set is trained with seed 0, but we test on ProGAN models with seed 1-9. For cross-loss testing, we test on CramerGAN~\cite{bellemare2017cramer} and SSGAN~\cite{chen2019ssgan} models, which have the same generator architecture as MMDGAN and InfoMaxGAN respectively but with different loss functions. Note that in cross-seed and cross-loss testing,  models are trained on the same dataset as models in the train-set to control the dataset variable. For cross-finetune testing, we test on models finetuned on the models in the train-set. We finetune with FFHQ-elders and LSUN-sofa respectively in the celebA and LSUN-bedroom experiment. For cross-dataset testing, we test on models trained on different datasets, e.g., in the celebA experiment, the models in the train-set are all trained on celebA, but we test on models all trained on LSUN-bedroom. All of the GAN images and real images in this dataset are 128px. 


\noindent{\bf Results.} The results are shown in Table~\ref{tab:exp1}, which are measured by accuracy. Compare DNA-Det (the last row) with existing methods (first three rows), we have several findings: 1) In closed-set testing, nearly all methods achieve relatively good performance, suggesting that features captured by these methods are sufficient for model-level attribution. 
2) In cross-testings, the performance degrades across all methods with different degrees. Among these cross-testings, the performance drops the most in cross-finetune and cross-dataset testing, showing that attribution methods are likely to learn content-relevant features, which is harmful for architecture attribution.
3) Compared with existing methods, DNA-Det achieves superior performance in closed-set and all cross-testings, especially gaining large improvement in cross-finetune and cross-dataset testing (from $\sim$ 30\% accuracy to over 80\%). As a result, DNA-Det is qualified for deepfake network architecture attribution. 

\begin{table}
\begin{center}
\scalebox{0.78}{
\begin{tabular}{lcccccc}
\toprule
GAN & Block Type & Skip Connect & Upsample & Norm \\
\midrule
ProGAN & DCGAN  & - & Nearest & PN \\
MMDGAN & ResNet & Upsample+Conv & Depth2Space&  BN \\ 
SNGAN & ResNet & Upsample+Conv  & Nearest & BN \\ 
InfoMaxGAN & ResNet  & Upsample+Conv  & Bilinear & BN\\
\bottomrule
\end{tabular}}
\end{center}
\caption{Structure components of four GANs.} 
\label{tab:component}
\end{table}

\begin{figure}[t]
\begin{center}
\includegraphics[width=\linewidth]{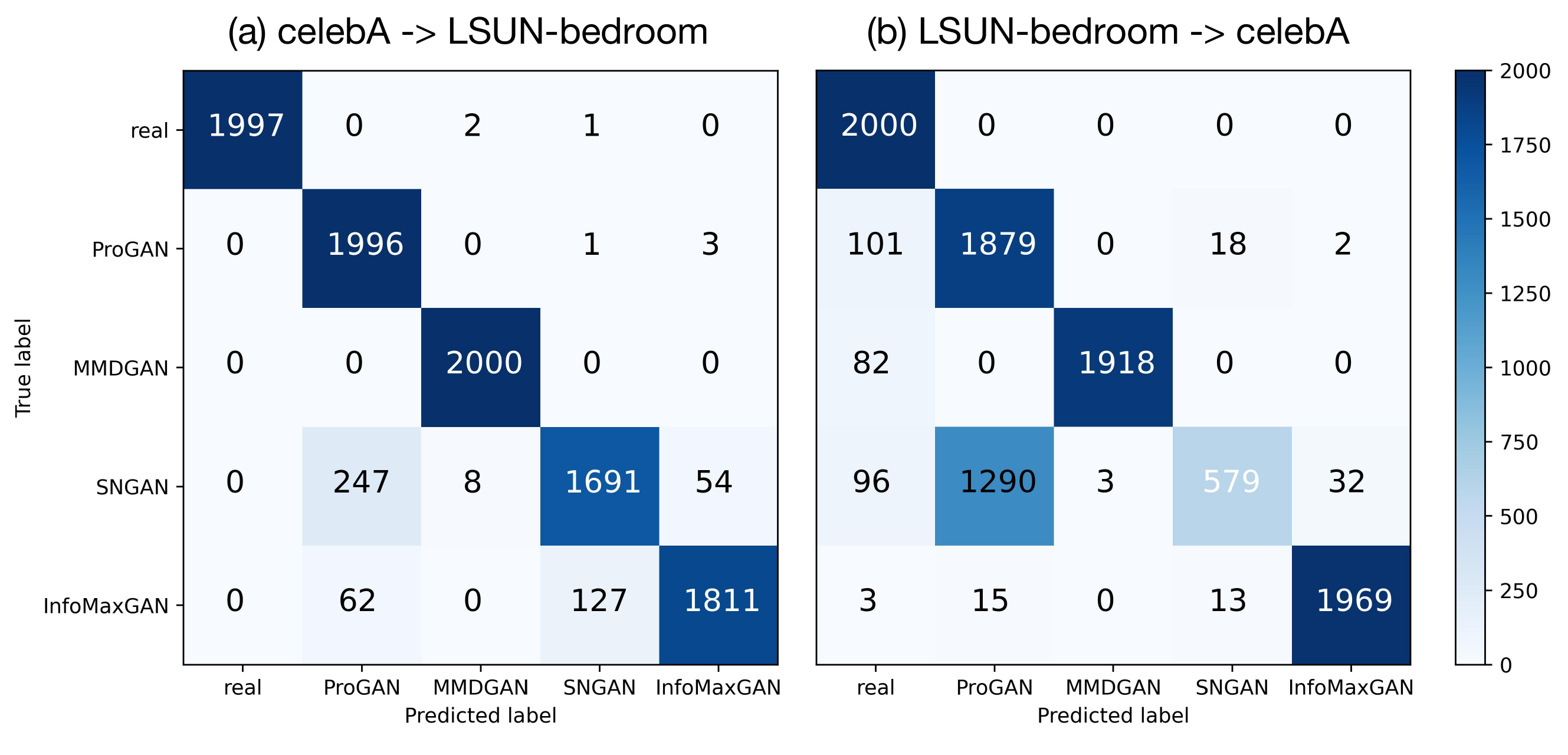} 
\end{center}
  \caption{Confusion matrices of two cross-dataset testings.} 
\label{fig:cm}
\end{figure}


\noindent{\bf Further Analysis.} We show the confusion matrices on the two cross-dataset testings in Figure~\ref{fig:cm}. From the confusion matrices, we find that SNGAN and InfoMaxGAN, SNGAN and ProGAN tend to be confused. To explore the reason, we check the details of the architecture components in Table~\ref{tab:component}. We notice that ProGAN and SNGAN both use Nearest upsampling layer but with different block structures (block type and skip connection). SNGAN and InfoMaxGAN share the same block structure but use different upsampling layers. From the relationship between the confusion matrices and architecture components, we have the following findings: 1) The successfully classified samples on the diagonal reflect that different block structures and upsampling types leave distinct traces, such that ProGAN and SNGAN, SNGAN and InfoMaxGAN can be distinguished in cross-dataset testing. 2) The misclassified samples suggest that the network doesn't capture the overall architecture traces on several samples, which causes the confusion between architectures whose components are partly the same.

\subsection{Evaluation on GANs in the Wild} 
\label{sec:q2}
\noindent{\bf Datasets.} In the real-world scenario, the collected data for different architectures may be more complex. The models may generate diverse contents and don not overlap among architectures. The content bias will mislead the network to focus on useless semantics. Thus we simulate the challenging real-world scenario and construct a large-scale dataset containing multiple public-released GANs with diverse contents as shown in Table~\ref{tab:multigan}. The dataset includes 59 GAN models from 10 architectures with 3 resolutions. Apart from the GANs used in Section~\ref{sec:q1}, we further include CycleGAN~\cite{zhu2017unpaired}, StackGAN2~\cite{stackganv2}, StyleGAN~\cite{karras2019style} and StyleGAN2~\cite{karras2020stylegan2}. Note that we take the different resolution versions of the same algorithm as different architectures, because they are different in the number of layers. The performance is measured by accuracy and macro-averaged F1-score over all classes. \\
\noindent{\bf Results.} 
From the results in Table~\ref{tab:exp2}, we can observe that: 1) With more GANs added, the experiment becomes more difficult as the accuracies of compared methods are all below 90\% in closed-set; 2) Our method outperforms other methods, not only in closed-set but also in cross-dataset testing, showing the effectiveness of our method in distinguishing different architectures and the generalization ability in real-world fake image architecture attribution.

\subsection{Ablation Study}
\label{sec:q3}
\begin{figure}[t]
\begin{center}
\includegraphics[width=\linewidth]{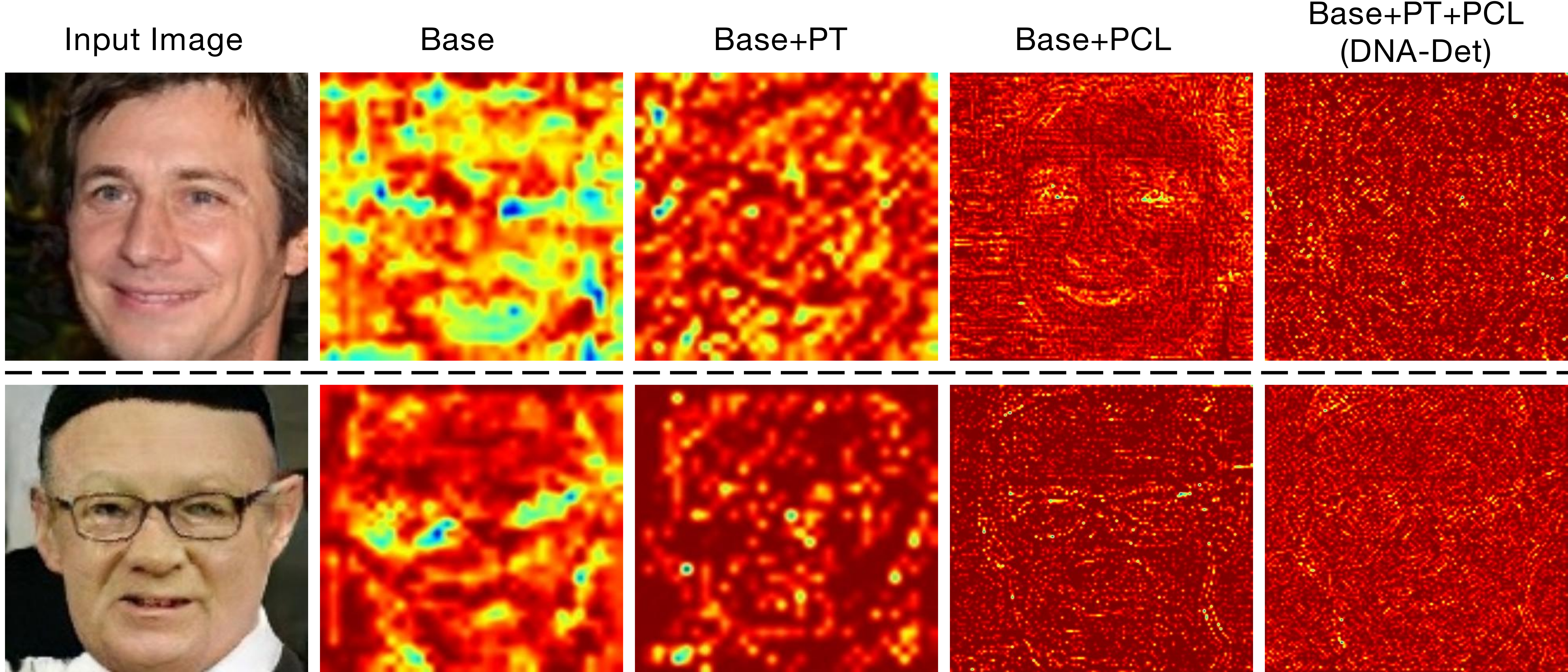} 
\end{center}
  \caption{Qualitative Analysis. Comparison of GradCAM heatmaps. Bluer color indicates a higher response for better visualization.} 
\label{fig:ablation}
\end{figure}

\noindent{\bf Quantitative Analysis.} The results in Table~\ref{tab:exp1} and Table~\ref{tab:exp2} validate the effectiveness of pre-train on image transformations (PT) and patchwise contrastive learning (PCL). Removing any of them on DNA-Det causes the performance to drop in nearly all settings. PCL is by far the most important one. In the hardest cross-dataset evaluation, removing it results in a dramatic drop of $41.95$, $33.72$ and $32.65$ points. This shows that the global consistency assumption makes sense and plays an important role in our method. 
Without PT, the performance drops by a modest $5.02$, $2.36$ and $3.28$ points, respectively. But when PT is added to the base network, it improves $19.98$, $28.07$ and $26.09$ points, which means features extracted by image transformation classification is related to architecture traces in some aspects. We also apply PCL and PT to the compared method AttNet, both results in a large improvement. The former improves $51.97$ and $58.18$ points, and the latter improves $6.17$ and $27.94$ in cross-dataset evaluation as shown in Table~\ref{tab:exp1}.\\
\noindent{\bf Qualitative Analysis.} We show in Figure~\ref{fig:ablation} the GradCAM~\cite{selvaraju2017grad} heatmaps to visualize how focused regions change with PCL and PT added. The two input images are from ProGAN and SNGAN respectively. The base network tends to concentrate on semantic-related local regions such as eyes and facial outline, which is untransferable for architecture attribution. With the PT added, the areas of concern are no longer locally focused. Adding PCL to the base network makes the feature extractor rely on more global and fine-grained traces, yet some salient regions such as eyes and face boundary can still be observed. PT plus PCL promote the network to only focus on globally consistent traces all around the image, and semantic-related regions nearly disappear in the heatmap. 

\begin{table}[t]
\begin{center}
\scalebox{0.85}{
\begin{tabular}{lcccc}
\toprule
\multirow{2}{*}{Method} & \multicolumn{2}{c}{closed-set} & \multicolumn{2}{c}{cross-dataset} \\
\cmidrule{2-5}
& Acc.& F1  & Acc. & F1\\
\midrule
PRNU(MIPR2019) & 66.77 	&63.76  & 20.31 &	12.58   \\ 
LeveFreq(ICML2020) & 70.53 &	73.71 	&	38.96 &	22.73  \\ 
AttNet(ICCV2019) & 84.57 &	86.48  &	53.21 &	33.14   \\ 
\midrule
Base & 88.11 &	90.27 &	47.95 &	25.04   \\
Base+PT &95.79 &	97.09& 73.02 &	50.82  \\
Base+PCL & \bf 99.99 &	\bf99.99  &	92.60 &	80.54  \\
\textbf{Base+PT+PCL(DNA-Det)} &99.96 &	99.98 &	\bf 92.94 &	\bf83.54  \\ 
\bottomrule
\end{tabular}}
\end{center}
\caption{Evalution on GANs in the wild for Section~\ref{sec:q2}}
\label{tab:exp2}
\end{table}

\begin{table}[t]
\begin{center}
\scalebox{0.75}{
\begin{tabular}{lcccccc}
\toprule
Method  & Crop & Blur   & JPEG  &  Noise & Relight & Combination\\
\midrule 
\bf {\textit{Closed-Set}}\\
PRNU &49.55	&71.00	&64.57	&72.77	&72.09	&40.71 \\
LeveFreq & 85.96	&77.58	&71.00	&85.07	&66.41	&45.37\\
AttNet &89.99	&89.76	&85.26	&90.65	&80.39	&73.40\\
\textbf{DNA-Det} &\bf 100.00 	&\bf 100.00 &	\bf 97.68 &	\bf 100.00 & \bf 96.39 &	\bf 80.16 \\
\midrule 
\bf {\textit{Cross-Dataset}} \\
PRNU & 20.29	&20.1&	20.58&	19.55&	19.09&	20.17\\
LeveFreq & 32.18&	30.67&	29.14&	31.70&	30.32&	23.83\\
AttNet & 24.01&	22.65&	23.89&	24.67&	23.70&	22.65\\
\textbf{DNA-Det} &\bf 82.48&\bf	82.69&\bf	76.43&\bf	81.72&\bf	78.90&\bf	59.53\\
\bottomrule
\end{tabular}}
\end{center}
\caption{Robustness analysis against common attacks.}
\label{tab:robust}
\end{table}

\begin{table}[t]
\begin{center}
\scalebox{0.75}{
\begin{tabular}{p{1.5cm}p{2cm}p{2cm}p{4cm}}
\toprule  
Resolution & Real, GAN & Train Content & Test Content\\
\midrule  
\multirow{5}{*}{128} & Real & celebA & bedroom \\
& ProGAN & celebA & bedroom   \\
& MMDGAN & celebA & bedroom \\
& SNGAN & celebA &  bedroom  \\
& InfoMaxGAN & celebA & bedroom \\
\midrule
\multirow{5}{*}{256} 
& Real & cat, airplane & boat, horse, sofa, cow, dog, train, bicycle, bottle, diningtable, motorbike, sheep, tvmonitor, bird, bus, chair, person, pottedplant, car \\
& ProGAN & cat, airplane & boat, horse, sofa, cow, dog, train, bicycle, bottle, diningtable, motorbike, sheep, tvmonitor, bird, bus, chair, person, pottedplant, car \\
& StackGAN2& cat, church & bird, bedroom, dog \\
& CycleGAN  & winter, orange & apple, horse, summer, zebra \\
& StyleGAN2 & cat, church & horse\\
\midrule
\multirow{3}{*}{1024} 
& Real & FFHQ & celebA-HQ \\
& StyleGAN & FFHQ & celebA-HQ, Yellow, Model, Asian Star, kid, elder, adult, glass, male, female, smile \\
& StyleGAN2 & FFHQ & Yellow, Wanghong, Asian Star, kid \\
\bottomrule 
\end{tabular}}
\end{center}
\caption{The dataset for Section~\ref{sec:q2}. For cross-dataset testing, the split makes sure training and testing models of each architecture don't overlap in content. 
}
\label{tab:multigan}
\end{table}

\subsection{Robustness Analysis}
We consider five types of attacks that perturb test images: noise, blur, cropping, JPEG compression, relighting and random combination of them. Detailed parameters of these attacks are the same with the work in~\cite{yu2019attributing}. Table~\ref{tab:robust} reports the closed-set and cross-dataset testing accuracy in the \textit{\textbf{celebA}} experiment under these attacks, which are included as a data augmentation in training for all methods. From the results, in closed-set testing, our method overcomes all attacks when any single attack is applied, and over other methods. The performance drops the most on combination attacks due to its complexity,  but we can still get an acceptable $80\%$ accuracy. In cross-dataset testing, our method can get almost $80\%$ accuracy under any of these attacks and a $59.53\%$ accuracy under combination attack, much superior to compared methods.

\section{Conclusions}
In this work, we present the first study on deepfake network architecture attribution.
Our empirical study verifies the existence of GAN architecture fingerprints, which are globally consistent on GAN images. Based on the study, we develop a simple yet effective approach named by DNA-Det to capture architecture traces by adopting pre-training on image transformations and patchwise contrastive learning. We evaluate DNA-Det on multiple cross-test setups and a large-scale dataset including 59 models derived from 10 architectures, verifying DNA-Det's effectiveness.
\section{Acknowledgements}
The corresponding author is Juan Cao. The authors thank Qiang Sheng, Xiaoyue Mi, Yongchun Zhu and anonymous reviewers for their insightful comments. 
This work was supported by the Project of Chinese Academy of Sciences (E141020), the Project of Institute of Computing Technology, Chinese Academy of Sciences (E161020), Zhejiang Provincial Key Research and Development Program of China (No. 2021C01164), and the National Natural Science Foundation of China (No. 62172420).

{\small
\bibliography{aaai22}
}

\setcounter{secnumdepth}{0}

\section{Appendix}
\begin{figure*}[h]
\begin{center}
\includegraphics[width=\linewidth]{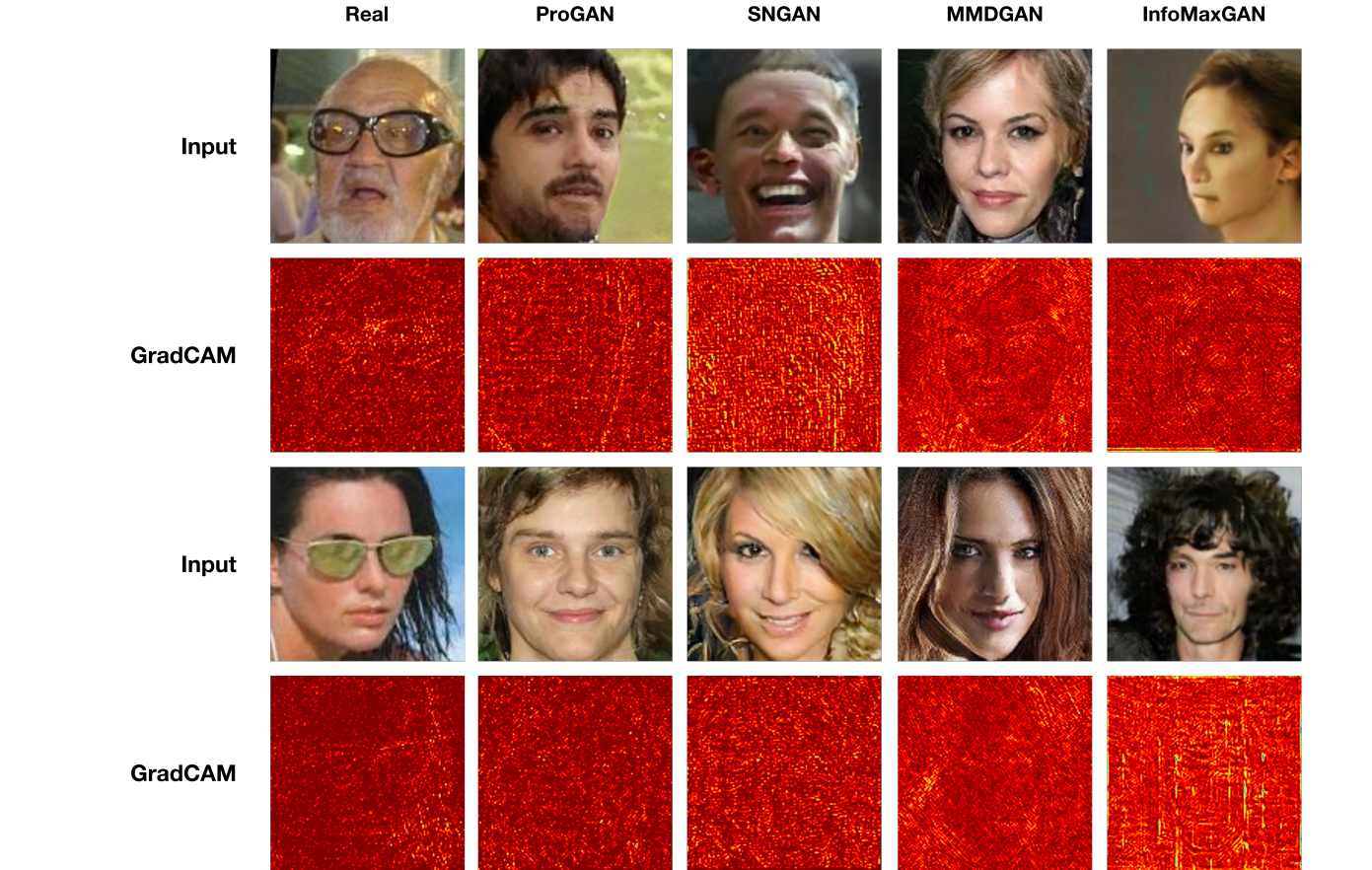} 
\end{center}
  \caption{More GradCAM visualization results. The results indicate that our network tends to focus on globally consistent traces.}
\label{fig:gradcam2}
\end{figure*}

The appendix is organized as follows:
\begin{itemize}
\setlength{\itemsep}{0pt}
\setlength{\parsep}{0pt}
\setlength{\parskip}{0pt}
\item 
Section A gives the architecture details of DNA-Det.
\item 
Section B includes detailed architecture descriptions for four GANs in Section~\ref{sec:q1}.
\item
Section C compares the convergence speed with and without pre-training on image transformations.
\item
Section D shows the testing accuracies for multiple cross-test setups in step2's different training epochs. 
\item
Section E provides ablation study on resize size and patch crop ratio to the cross-dataset accuracy.
\item
Section F includes complete experiment results for the empirical study in Section~\ref{sec:study}.
\item
Figure~\ref{fig:gradcam2} shows more GradCAM visualization results.
\end{itemize}

\section{A. DNA-Det Architecture}
\label{sec:A}
DNA-Det's architecture details are described in Table~\ref{tab:dna-det}, which consists of an encoder, a classification head and a projection head. The Kernel column stands for the kernel size of convolutional layers, which is described in format $[filter\_h, filter\_w, stride]$. LReLU stands for Leaky ReLU and the slope for all LReLU functions is set to 0.2. 
\begin{table}[h]
\begin{center}
\scalebox{0.8}{
\begin{tabular}{llcc}
\toprule
& Layer & Kernel  & Output \\
\midrule
\multirow{10}{*}{Encoder} &Input &-& 64$\times$64$\times$3 \\
 &Conv, BN, LReLU & [4,4,2] & 32$\times$32$\times$64 \\
 &Conv, BN, LReLU & [3,3,1] & 32$\times$32$\times$64 \\
 &Conv, BN, LReLU & [4,4,2] & 16$\times$16$\times$128 \\
 &Conv, BN, LReLU & [3,3,1] & 16$\times$16$\times$128 \\
 &Conv, BN, LReLU & [4,4,2] & 8$\times$8$\times$256 \\
 &Conv, BN, LReLU & [3,3,1] & 8$\times$8$\times$256 \\
 &Conv, BN, LReLU & [4,4,2] & 4$\times$4$\times$512 \\
 &Conv, BN, LReLU & [3,3,1] & 4$\times$4$\times$512 \\
 &AvgPool & - & 512 \\
\midrule
\multirow{4}{*}{Projection head} & Linear & -& 512\\
& ReLU & - & 512 \\
& Linear & -& 128 \\
& Normalize & - & 128 \\
\midrule
Classfication head&Linear &- & class num\\
\bottomrule
\end{tabular}}
\end{center}
\caption{DNA-Det architecture} 
\label{tab:dna-det}
\end{table}

\section{B. Architecture Details for GANs}
\label{sec:B}
The architectures of ProGAN, MMDGAN, SNGAN and InfoMaxGAN are described in Table~\ref{tab:progan}, Table~\ref{tab:mmdgan}, Table~\ref{tab:sngan} and Table~\ref{tab:infomaxgan}. The generator's input is a 1-d latent code named $z$. The RS column stands for the resample layer. BN stands for batch normalization and PN stands for pixel normalization. ResBlock stands for the generation block with a ResNet-like architecture. In ResBlock, two paths are merged by an addition layer. DCBlock stands for the generation block with a DCGAN-like architecture. $h$ and $w$ are the input shape to the generation block. $c_{i}$ and $c_{o}$ are the input channels and output channels for a block. 
From the architectures, we can find that SNGAN and ProGAN both use Nearest upsampling layer but with different blocks. SNGAN and InfoMaxGAN share the same block structure but use different upsampling layers.

\section{C. Convergence Speed Comparison}
\label{sec:D}

\begin{figure}[t]
\begin{center}
\includegraphics[width=\linewidth]{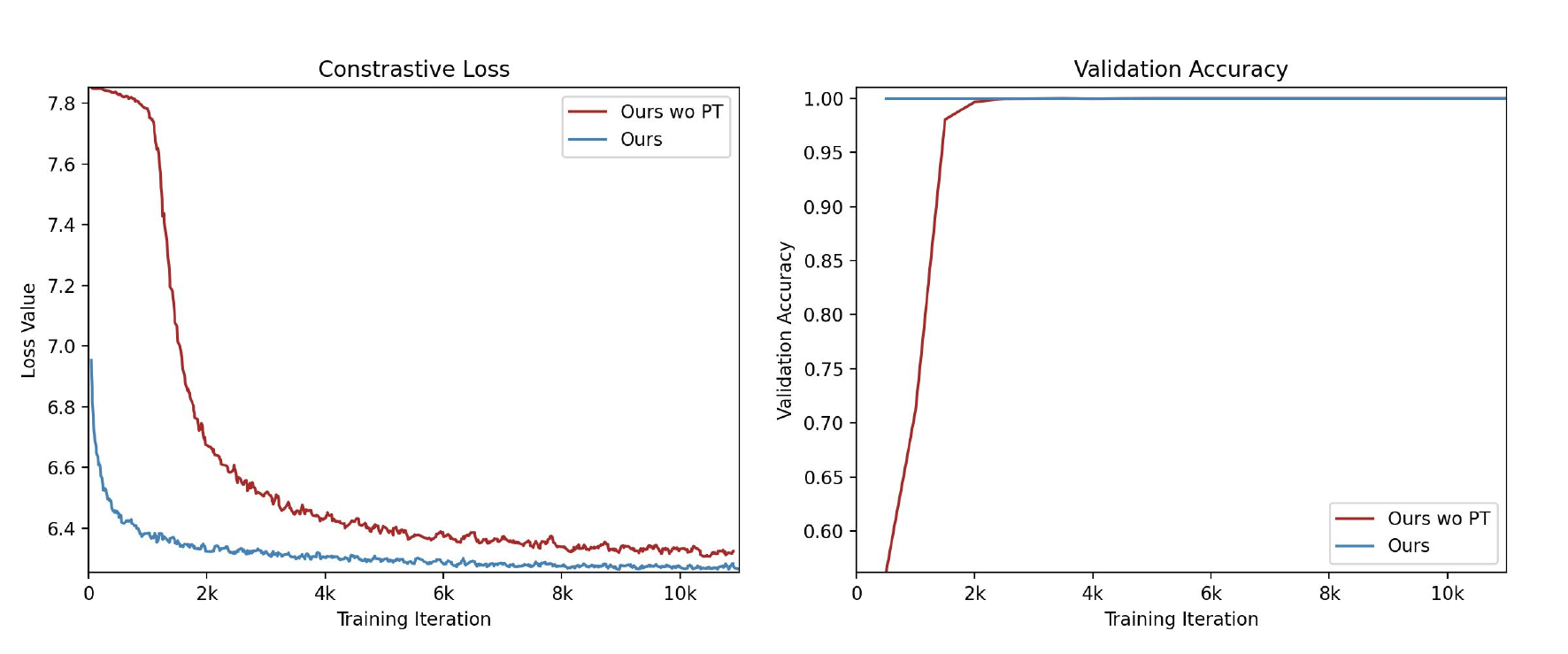} 
\end{center}
  \caption{Contrastive loss and val accuracy with and without pre-training on image transformations.} 
\label{fig:curve}
\end{figure}

We find pre-training on image transformations (PT) largely improves the convergence speed. We plot in Figure~\ref{fig:curve} the contrastive loss and validation accuracy in the \textit{\textbf{LSUN-bedroom}} experiment during training for our method with and without PT. It shows that PT results in much faster convergence, which indicates that PT could help the network to focus on globally consistent traces useful for architecture representation at the beginning of training.


\section{D. Testing Accuracy vs. Training Epochs}
\begin{figure}[h]
\begin{center}
\includegraphics[width=\linewidth]{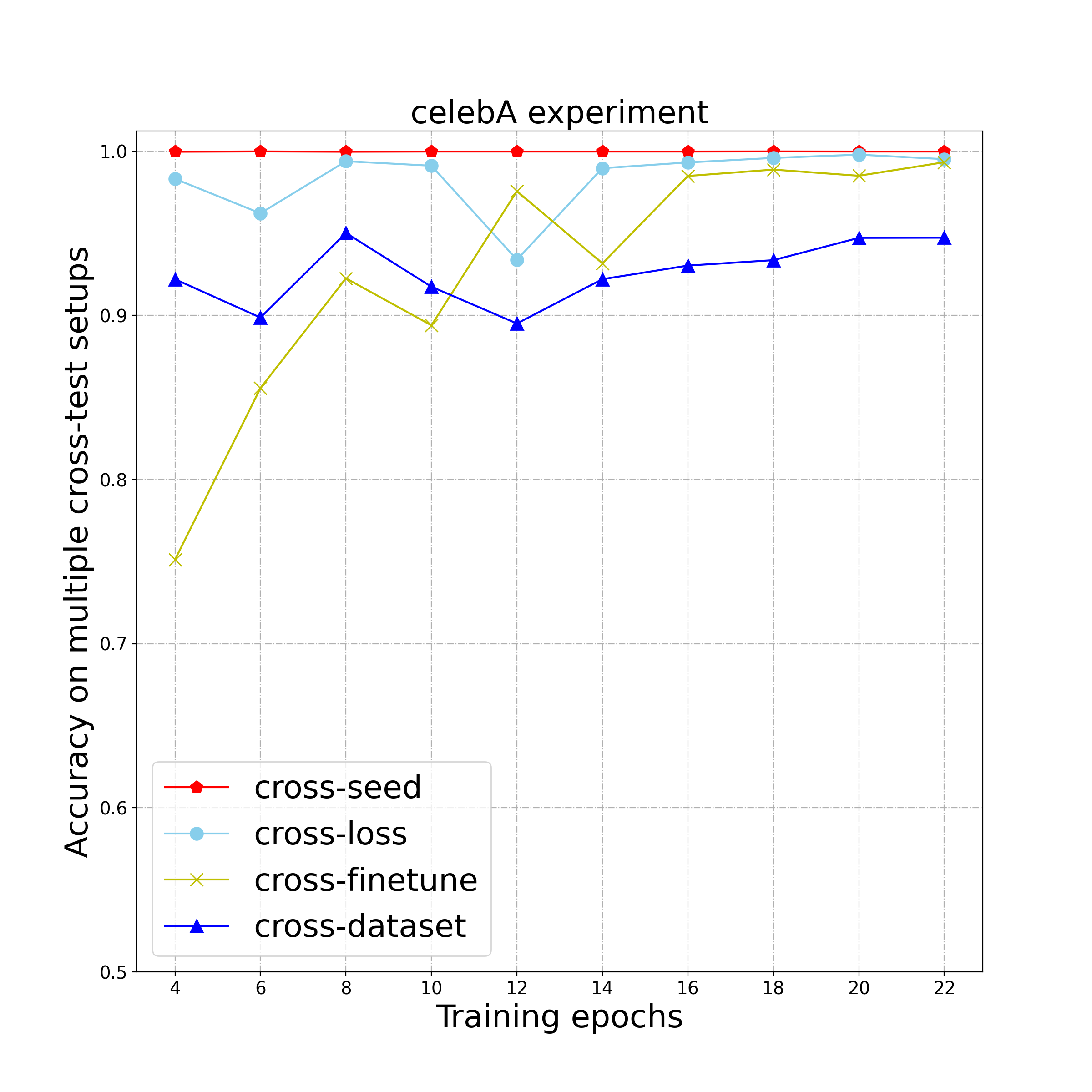} 
\end{center}
  \caption{Accuracy on multiple cross-test setups vs. training epochs} 
\label{fig:acc_curve}
\end{figure}

We show in Figure~\ref{fig:acc_curve} the multiple cross-test setups' accuracies in step2's different training epochs. As the figure shows, the results are relatively stable after a certain number of epochs. Thus we select all models in epoch 20 to produce the results.
 
\section{E. Effect of Resize Size and Patch Crop Ratio}
We preprocess CelebA and LSUN-bedroom images to $128\times128$. For CelebA dataset, we crop each image centered at (x,y) = (89,121) with size $128\times128$ before training. For LSUN-bedroom images, we center-crop them with size $128\times128$.

We find resizing images to a larger resolution magnifies the fingerprint, enhancing the performance of PCL, thus we resize images to 512px. For experiment in section~\ref{sec:q1}, we directly resize images to 512px and crop 64px patches as input. 
While for experiment in section~\ref{sec:q2}, images are different in resolution and direct resizing them to the same resolution will introduce resolution bias, e.g., a 128px image will be upsampled while a 1024px image will be downscaled. Thus, we firstly rescale them to an equal 128px size (the lowest resolution in the dataset) before resizing to 512px. We don't crop them to 128px, as it will introduce content bias, e.g., a 128px patch from 1024px GAN face image only contains a small part of the face, but the entire face for a 128px GAN. 

We provide an ablation study on resize size and patch crop ratio for PCL in Table~\ref{tab:size_ablation}, which shows a 512px resize size and 1/8×1/8 crop ratio (64px) is the best.   

\begin{table}[t]
\begin{center}
\scalebox{0.95}{
\begin{tabular}{lccccc}
\toprule
\multirow{2}{*}{Resize size} & \multicolumn{5}{c}{Patch crop ratio} \\
\cmidrule{2-6}
& $1^{2}$ & ${1/2}^{2}$ & ${1/4}^{2}$ & ${1/8}^{2}$ & ${1/16}^{2}$ \\
\midrule
128 & 53.00 & 49.42 & 54.32 & 21.04 & - \\
256 & 52.80 & 70.34 & 71.45 & 73.88 & 31.57 \\
384 & 56.35 & 82.38 & 90.54 & 78.52 & 45.30 \\
512 & 65.95 & 74.61 & 87.01 & \textbf{94.95} & 63.42 \\
640 & 52.29 & 71.17 & 88.28 & 87.94 & 71.08 \\
\bottomrule
\end{tabular}}
\end{center}
\caption{Ablation study on resize size and patch crop ratio}
\label{tab:size_ablation}
\end{table}

\begin{table*}[h]
\begin{subtable}{0.5\linewidth}
\begin{center}
\begin{tabular}{lccc}
\toprule
Layer & Kernel & Output \\
\midrule
z & - & 512 \\
PN & - & 512\\
Linear,Reshape & - &  8192 $\times$512 \\ 
LReLU,PN & -&  4 $\times$ 4 $\times$ 512 \\
Conv & [3,3,1] & 4  $\times$ 4 $\times$ 512 \\ 
LReLU,PN & -  &   4$\times$ 4 $\times$512 \\
DCBlock & [3,3,1]  & 8 $\times$ 8 $\times$ 512 \\
DCBlock & [3,3,1] & 16 $\times$ 16 $\times$ 512 \\
DCBlock & [3,3,1]  & 32 $\times$ 32 $\times$ 512 \\
DCBlock & [3,3,1]& 64 $\times$ 64 $\times$ 256 \\
DCBlock & [3,3,1] & 128 $\times$ 128 $\times$ 128 \\
Conv & [3,3,1] & 128 $\times$ 128 $\times$ 3 \\
\bottomrule
\end{tabular}
\end{center}
\caption{ProGAN generator} 
\end{subtable}
\begin{subtable}{0.5\linewidth}  
\begin{center}
\begin{tabular}{lcccc}
\toprule
Layer & Kernel & RS & Output \\
\midrule
Upsample & - & Nearest UP & $2h \times 2w \times c_{i}$ \\
Conv & [3,3,1] & - & $2h \times 2w \times c_{o}$ \\
LReLU,PN & - & - & $2h \times 2w \times c_{o}$ \\
Conv & [3,3,1] & - & $2h \times 2w \times c_{o}$ \\
LReLU,PN & - & - & $2h \times 2w \times c_{o}$ \\
\bottomrule
\end{tabular}
\end{center}
\caption{ProGAN generator DCBlock}
\end{subtable}
\caption{ProGAN architecture}
\label{tab:progan}
\end{table*}

\begin{table*}[h]
\begin{subtable}{0.5\linewidth}
\begin{center}
\begin{tabular}{lccc}
\toprule
Layer & Kernel  & Output \\
\midrule
z & -& 100 \\
Linear & - & 4$\times$4$\times$1024 \\
ResBlock & [3,3,1]& 8$\times$8$\times$512 \\
ResBlock & [3,3,1]& 16$\times$16$\times$256 \\
ResBlock & [3,3,1]& 32$\times$32$\times$128 \\
ResBlock & [3,3,1] & 64$\times$64$\times$64 \\
BN,ReLU & - & 64$\times$64$\times$64 \\
Deconv & [5,5,2] & 128$\times$128$\times$3 \\
Sigmoid & - &  128$\times$128$\times$3 \\
\bottomrule
\end{tabular}
\end{center}
\caption{MMDGAN generator} 
\end{subtable}
\begin{subtable}{0.5\linewidth}  
\begin{center}
\begin{tabular}{lcccc}
\toprule
Layer & Kernel & RS & Output \\
\midrule
Upsample & - & Depth-to-space Up & $2h \times 2w \times c_{i}$\\
Conv & [3,3,1] & - & $2h \times 2w \times c_{o}$\\
\midrule
BN,ReLU & - & - & $h \times w \times c_{i}$ \\
Upsample & - & Depth-to-space Up & $2h \times 2w \times c_{i}$\\
Conv & [3,3,1] & - & $2h \times 2w \times c_{o}$\\
BN,ReLU & - & - & $2h \times 2w \times c_{o}$ \\
Conv & [3,3,1] & - & $2h \times 2w \times c_{o}$\\
\midrule
Addition & - & - & $2h \times 2w \times c_{o}$ \\
\bottomrule
\end{tabular}
\end{center}
\caption{MMDGAN generator ResBlock}
\end{subtable}
\caption{MMDGAN architecture}
\label{tab:mmdgan}
\end{table*}

\begin{table*}[h]
\begin{subtable}{0.5\linewidth}
\begin{center}
\begin{tabular}{lccc}
\toprule
Layer & Kernel &Output \\
\midrule
z & - & 128 \\
Linear & - & 4$\times$4$\times$1024 \\
ResBlock & [3,3,1]  & 8$\times$8$\times$1024 \\
ResBlock & [3,3,1]  & 16$\times$16$\times$512 \\
ResBlock & [3,3,1]  & 32$\times$32$\times$256 \\
ResBlock & [3,3,1]  & 64$\times$64$\times$128 \\
ResBlock & [3,3,1]  & 128$\times$128$\times$64 \\
BN,ReLU & - & 128$\times$128$\times$64 \\
Conv & [3,3,1]  & 128$\times$128$\times$3 \\
Tanh & - &  128$\times$128$\times$3 \\
\bottomrule
\end{tabular}
\end{center}
\caption{SNGAN generator} 
\end{subtable}
\begin{subtable}{0.5\linewidth}  
\begin{center}
\begin{tabular}{lcccc}
\toprule
Layer & Kernel & RS & Output \\
\midrule
Upsample & - & Nearest Up & $2h \times 2w \times c_{i}$\\
Conv & [3,3,1] & - & $2h \times 2w \times c_{o}$\\ 
\midrule
BN,ReLU & - & - & $h \times w \times c_{i}$ \\
Upsample & - & Nearest Up & $2h \times 2w \times c_{i}$\\
Conv & [3,3,1] & - & $2h \times 2w \times c_{o}$\\
BN,ReLU & - & - & $2h \times 2w \times c_{o}$ \\
Conv & [3,3,1] & - & $2h \times 2w \times c_{o}$\\
\midrule
Addition & - & - & $2h \times 2w \times c_{o}$ \\
\bottomrule
\end{tabular}
\end{center}
\caption{SNGAN generator ResBlock} 
\end{subtable}
\caption{SNGAN architecture}
\label{tab:sngan}
\end{table*}

\begin{table*}[h]
\begin{subtable}{0.5\linewidth}
\begin{center}
\begin{tabular}{lccc}
\toprule
Layer & Kernel  & Output \\
\midrule
z & -  & 128 \\
Linear & -  & 4$\times$4$\times$1024 \\
ResBlock & [3,3,1]  & 8$\times$8$\times$1024 \\
ResBlock & [3,3,1]  & 16$\times$16$\times$512 \\
ResBlock & [3,3,1]  & 32$\times$32$\times$256 \\
ResBlock & [3,3,1]  & 64$\times$64$\times$128 \\
ResBlock & [3,3,1]  & 128$\times$128$\times$64 \\
BN,ReLU & -  & 128$\times$128$\times$64 \\
Conv & [3,3,1]  & 128$\times$128$\times$3 \\
Tanh & -  &  128$\times$128$\times$3 \\
\bottomrule
\end{tabular}
\caption{InfoMaxGAN generator} 
\end{center}
\end{subtable}
\begin{subtable}{0.5\linewidth}  
\begin{center}
\begin{tabular}{lcccc}
\toprule
Layer & Kernel & RS & Output \\
\midrule
Upsample & - & Bilinear Up & $2h \times 2w \times c_{i}$\\
Conv & [3,3,1] & - & $2h \times 2w \times c_{o}$\\ 
\midrule
BN,ReLU & - & - & $h \times w \times c_{i}$ \\
Upsample & - & Bilinear Up & $2h \times 2w \times c_{i}$\\
Conv & [3,3,1] & - & $2h \times 2w \times c_{o}$\\
BN,ReLU & - & - & $2h \times 2w \times c_{o}$ \\
Conv & [3,3,1] & - & $2h \times 2w \times c_{o}$\\
\midrule
Addition & - & - & $2h \times 2w \times c_{o}$ \\
\bottomrule
\end{tabular}
\caption{InfoMaxGAN generator ResBlock} 
\end{center}
\end{subtable}
\caption{InfoMaxGAN architecture}
\label{tab:infomaxgan}
\end{table*}

\section{F. More Results for Empirical Study}

In Section~\ref{sec:study}, we perform an empirical experiment to explore the property of architecture traces, and observe that architecture leaves global consistent traces while weights tend to leave local traces. In this section, we report the complete results of this experiment. 
The four GAN models for architecture classification are ProGAN, MMDGAN, SNGAN and InfoMaxGAN models trained on CelebA dataset. The four GAN models for weight classification are ProGAN models trained on celebA but with different training seeds. These models all generated images of 128px. The images are firstly resized to 512px and then cropped into patches of 128px according to the positions. We have tried two setups for the experiment. In setup-I, we use the base network with random initialization. In setup-II, we initialize the base network with the weights pre-trained on images transformations. The results that we report in Section~\ref{sec:study} come from the setup-II experiment.

\subsection{F.1. Setup-I}

\begin{figure*}[h]
\begin{center}
\includegraphics[width=\linewidth]{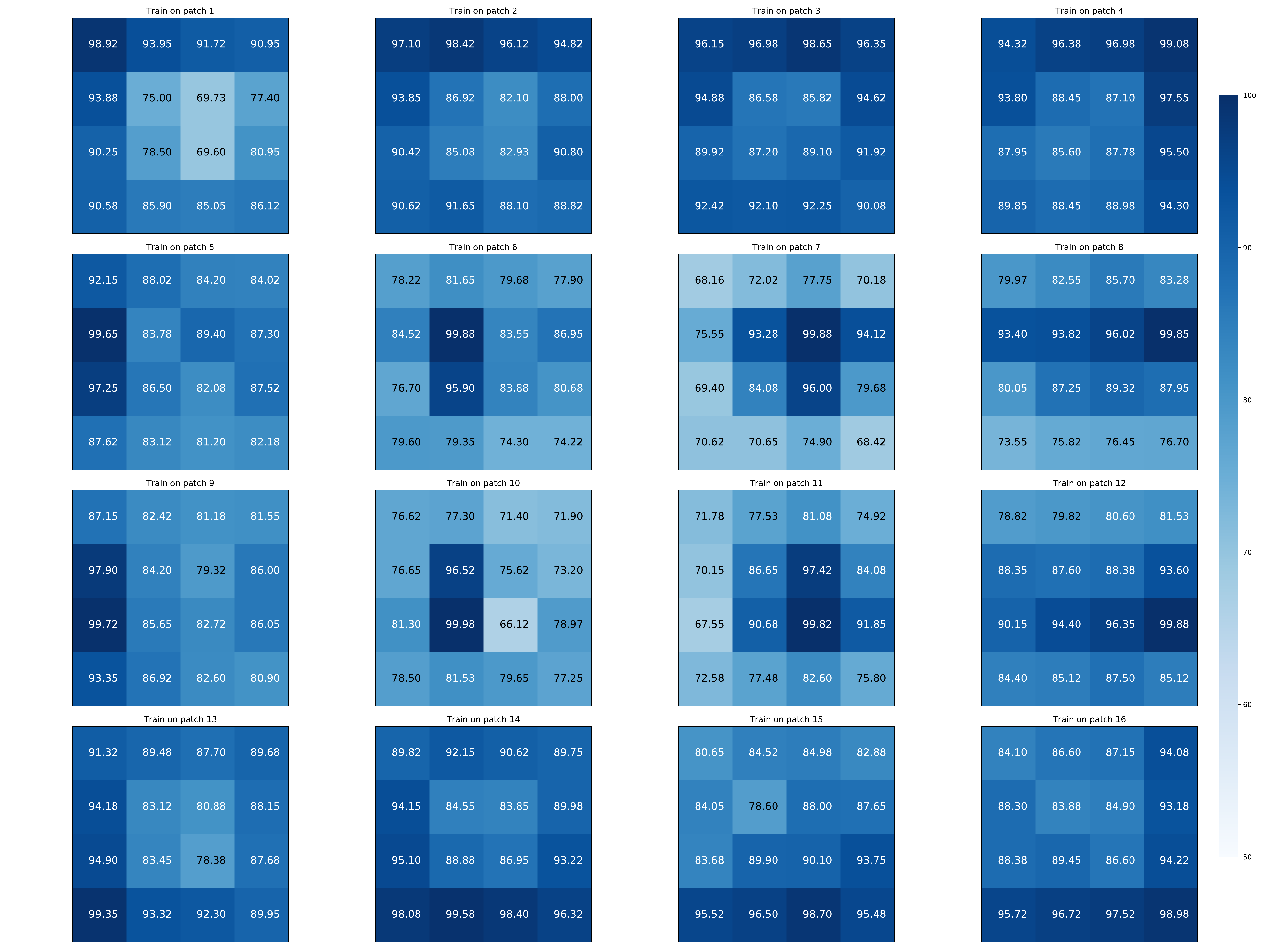} 
\end{center}
  \caption{Setup-I experiment results of empirical study for architecture classification.}
\label{fig:cm_4arch}
\end{figure*}

\begin{figure*}[h]
\begin{center}
\includegraphics[width=\linewidth]{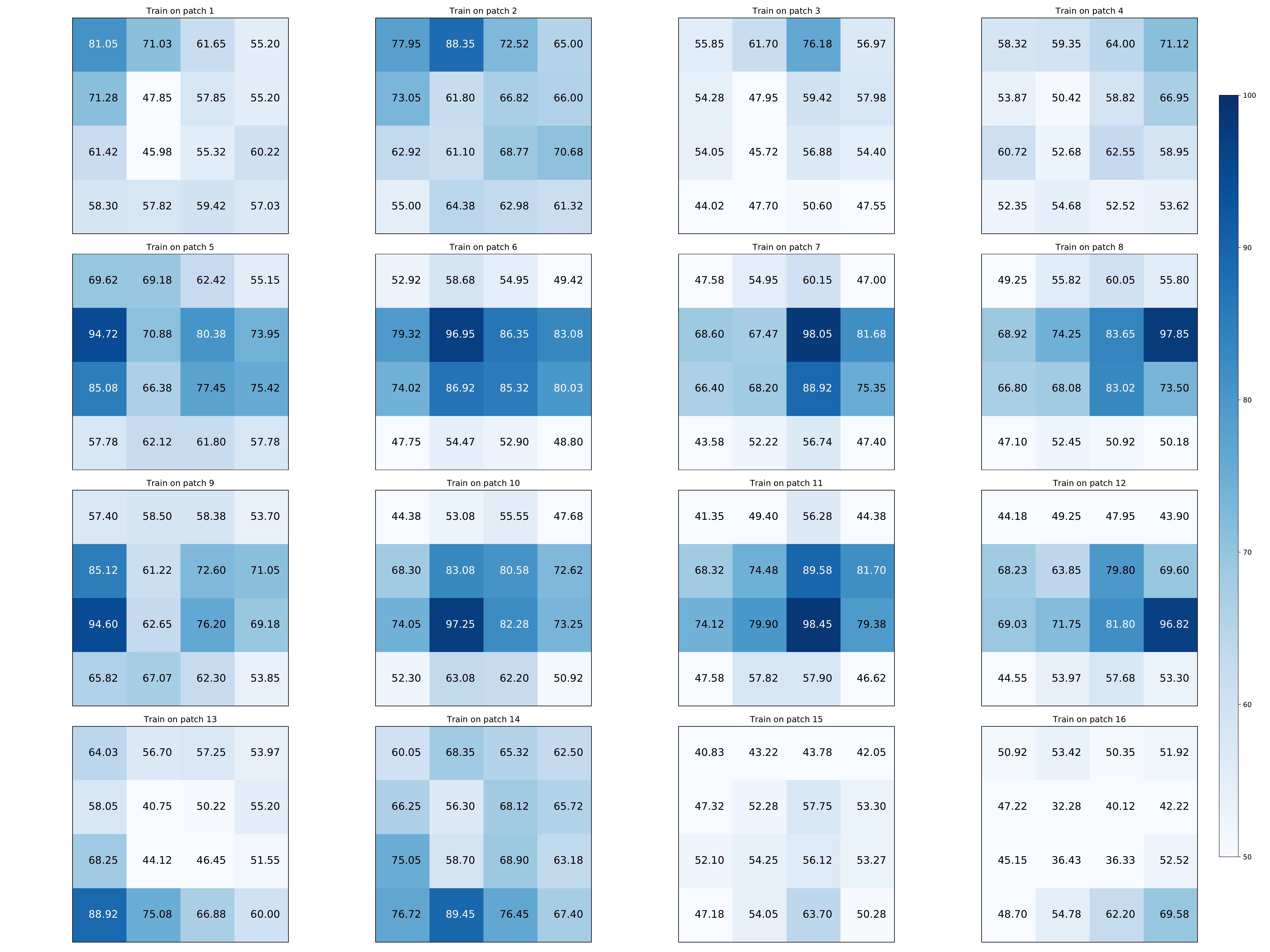} 
\end{center}
  \caption{Setup-I experiment results of empirical study for weight classification.} 
\label{fig:cm_4seed}
\end{figure*}

In setup-I, we use the base network without any further training strategy. We present the complete experiment results of architecture classification in Figure~\ref{fig:cm_4arch} and the weight classification results in Figure~\ref{fig:cm_4seed}. There are 16 subfigures in one figure, each represents an experiment trained on patches from a single position. 

In architecture classification, we can find that: 1) The testing accuracies are almost all higher than 70\% and most of them are higher than 80\% (numbered in white color) even though we train solely on patches from a single position. This indicates that there exist globally consistent traces that can discriminate models of different architectures. 2) The testing accuracy is highest in the position from which the patches are used for training. The reason may be that the base network is trained without any further strategies, it may have learned some local traces(maybe weight traces) other than architecture traces, such that the extracted traces from other positions are not consistent with the traces from training patches. To further check whether there really exists globally consistent traces in architecture classification, we enhance the network's ability to extract global traces in setup-II.

In weight classification, we can find that: 
1) There is a discrepancy in testing accuracies on different source positions (positions from which the patches are used for training). When training on patch 5-12, the testing accuracies on source positions all reach 90\%+. However, when training on patch 1-4 and 13-16, the testing accuracies on source positions are all below 90~\%, even less than 70~\% on patch 15 and patch 16. 
This phenomenon suggests that, weight traces are easier to be extracted in center regions, which may be because the amount of weight traces is different in different regions. 
2) There is another obvious discrepancy in testing accuracies on source positions and other positions. For example, when training on patch 5-12, the testing accuracies on source positions are as high as 94\%+, but drop at least 10\% and at most 40\% on other positions. This indicates that weight traces may vary in different regions.

\subsection{F.2. Setup-II}

\begin{figure*}[h]
\begin{center}
\includegraphics[width=\linewidth]{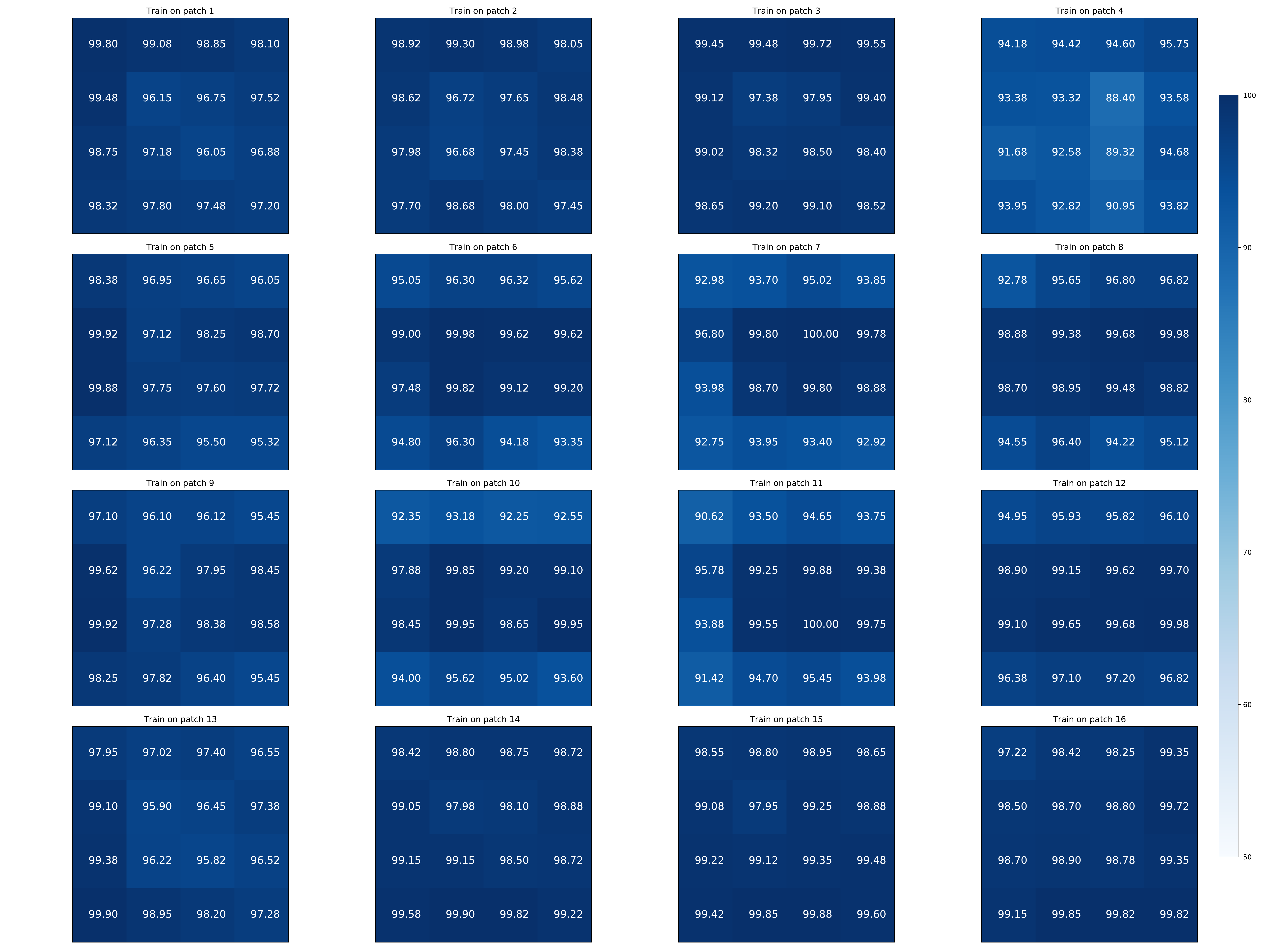} 
\end{center}
  \caption{Setup-II experiment results of empirical study for architecture classification.
  }
\label{fig:cm_4arch_ssl}
\end{figure*}

\begin{figure*}[h]
\begin{center}
\includegraphics[width=\linewidth]{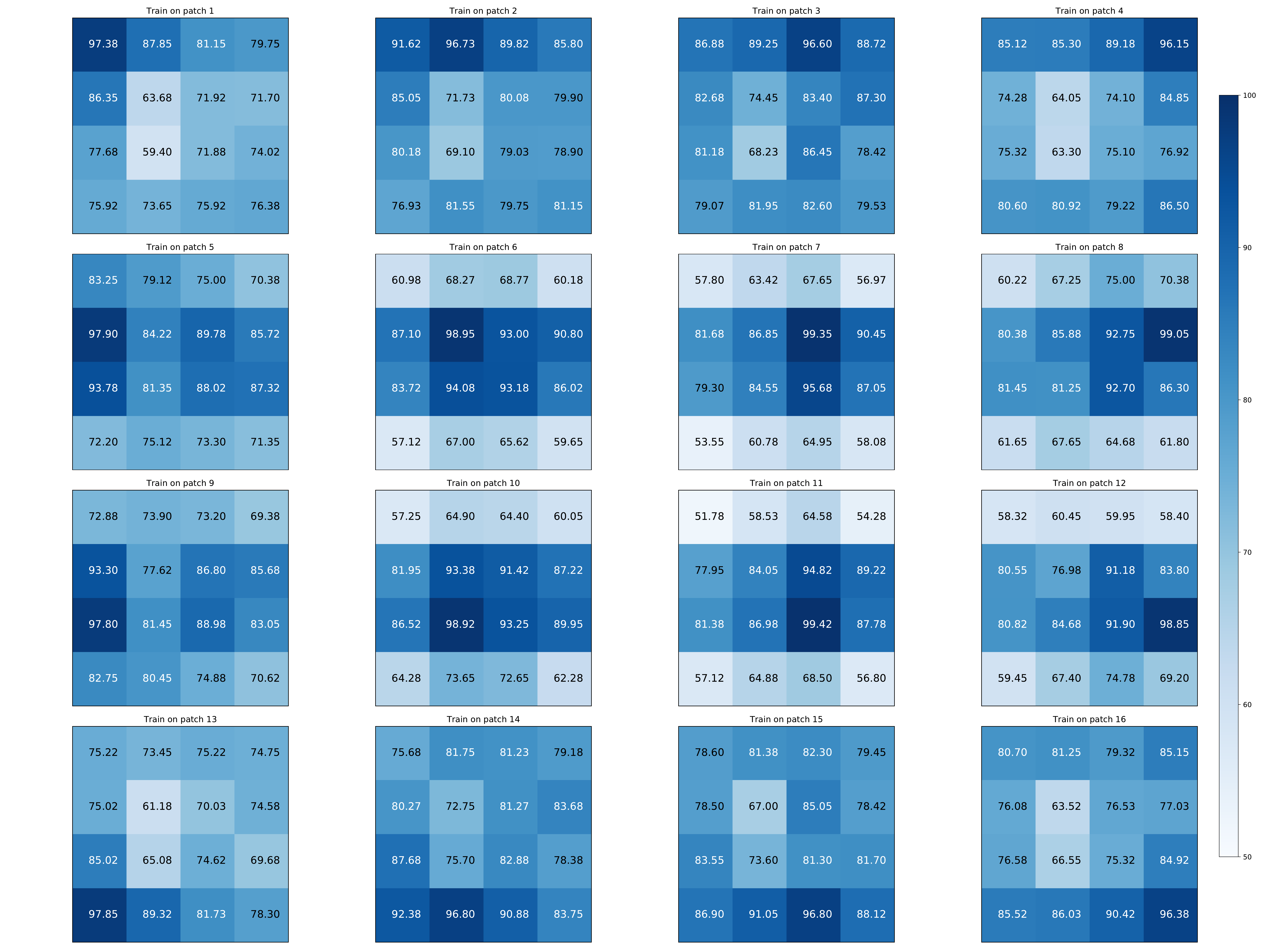} 
\end{center}
  \caption{Setup-II experiment results of empirical study for weight classification.}
\label{fig:cm_4seed_ssl}
\end{figure*}

It's still possible that if we utilize a more powerful network, which is good at extract global features, the results of weight classification would approach architecture classification.
To eliminate this uncertainty, we apply the strategy of pre-training on image transformations described in Section~\ref{sec:pretrain} to the base network, which would enhance the ability to extract globally consistent traces.
We name this experimental setup as setup-II. Figure~\ref{fig:cm_4arch_ssl} and Figure~\ref{fig:cm_4seed_ssl} show the setup-II results of architecture classification and weight classification respectively. 
We can find that: 1) The testing accuracies in architecture classification are almost all above 90\%, which indicates an apparent global consistency. 2) In weight classification, testing on the source position gets a nearly 100\% accuracy while not in setup-I, showing the pre-train strategy helps the network to extract subtle traces. However, the locally biased phenomenon is still remarkable in weight classification. For example, when training on patch 11, the accuracy is as high as 99\% on patch 11, but drops to around 50\% on many other patches.
These findings further confirm the conclusions in Setup-I. The results on patch 1 and 10 are selected to be presented in Section~\ref{sec:study}.

\end{document}